\documentclass[10pt,twocolumn,letterpaper]{article}

\usepackage[pagenumbers]{cvpr} 

\usepackage{siunitx}
\usepackage{tabularx}
\usepackage{booktabs}
\usepackage{pifont}
\usepackage[table]{xcolor}
\usepackage{circledsteps}
\usepackage{lipsum}
\usepackage[ruled, lined, linesnumbered, longend]{algorithm2e}

\newcommand{\cmark}{\ding{51}}%

\definecolor{pgreen}{rgb}{0.11, 0.62, 0.47}
\definecolor{ppurple}{rgb}{0.46, 0.44, 0.70}
\definecolor{pgrey}{rgb}{0.67, 0.67, 0.67}
\definecolor{porange}{rgb}{0.85, 0.36, 0.01}

\definecolor{cvprblue}{rgb}{0.21,0.49,0.74}
\usepackage[pagebackref,breaklinks,colorlinks,citecolor=cvprblue]{hyperref}

\def\paperID{} 
\def\confName{3DV\xspace}
\def\confYear{2024\xspace}

\title{Applying Plain Transformers to Real-World Point Clouds}

\author{Lanxiao Li \quad\quad Michael Heizmann \\
Karlsruhe Institute of Technology, Karlsruhe, Germany \\
{\tt\small lanxiao.li@kit.edu} \quad\quad {\tt\small michael.heizmann@kit.edu}}

\begin{document}
\maketitle

\begin{abstract}
To apply transformer-based models to point cloud understanding, many previous works modify the architecture of transformers by using, e.g., local attention and down-sampling. Although they have achieved promising results, earlier works on transformers for point clouds have two issues. First, the power of plain transformers is still under-explored. Second, they focus on simple and small point clouds instead of complex real-world ones. This work revisits the plain transformers in real-world point cloud understanding. We first take a closer look at some fundamental components of plain transformers, e.g., patchifier and positional embedding, for both efficiency and performance. To close the performance gap due to the lack of inductive bias and annotated data, we investigate self-supervised pre-training with masked autoencoder (MAE). Specifically, we propose drop patch, which prevents information leakage and significantly improves the effectiveness of MAE. Our models achieve SOTA results in semantic segmentation on the S3DIS dataset and object detection on the ScanNet dataset with lower computational costs. Our work provides a new baseline for future research on transformers for point clouds. 

\end{abstract}

\section{Introduction}
\label{sec:intro}
Transformers~\cite{vaswani2017transformer} have shown promising performance in computer vision tasks in recent years~\cite{dosovitskiy2020vit, liu2021swin}. The most representative model is Vision Transformer (ViT)~\cite{dosovitskiy2020vit}, which models an image as a sequence and extracts features using a \emph{plain} transformer encoder. It is called plain since ViT consists of stacked transformer layers and does not incorporate inductive biases, \eg, translation equivariance and locality, which are, on the contrary, essential ingredients in CNNs. Although simple and effective, a plain transformer requires more training data or careful design to achieve good performance in image processing~\cite{chen2021mocov3, dosovitskiy2020vit, xiao2021early}. 
\begin{figure}[t]
    \centering
    \includegraphics[width=\linewidth]{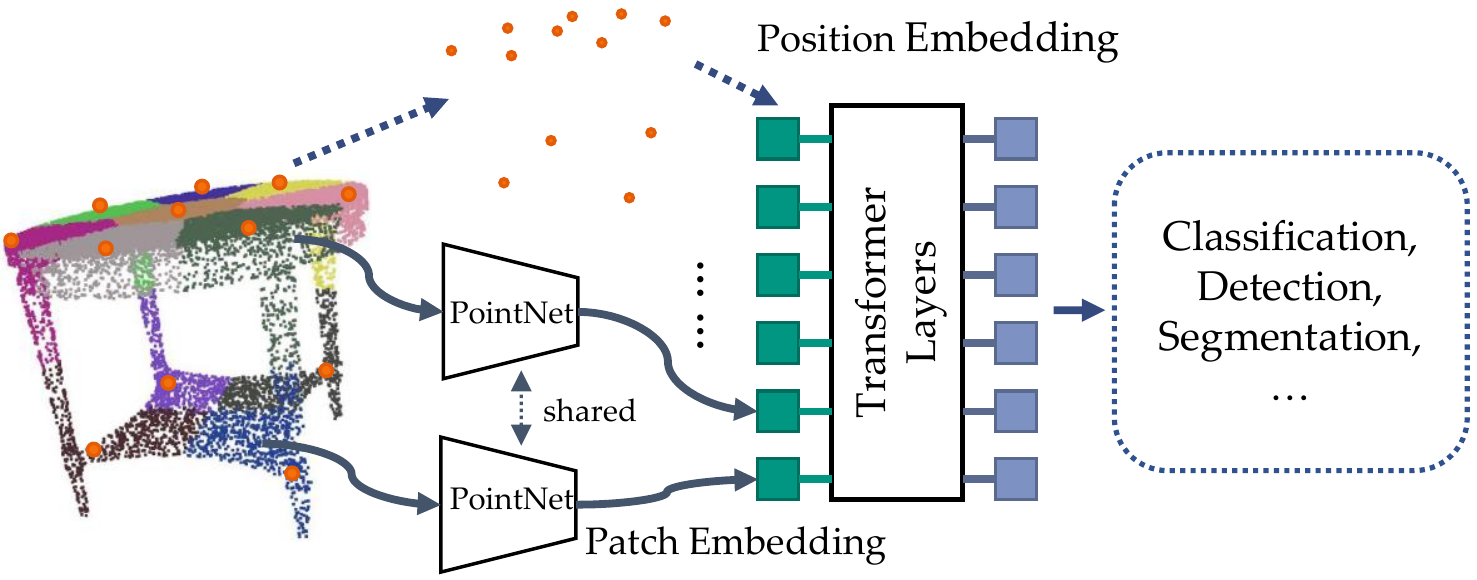}
    \caption{A plain transformer for point clouds. It simply uses stacked transformer layers without further modification. }
    \label{fig:transformer}
\end{figure}

Because of their capability to capture informative features, transformer-based methods are also attractive in 3D compuer vision. A lot of methods have been proposed to utilize transformers for point cloud understanding~\cite{engel2021point_transformer, guo2021pct, Hui2021pyrimidPCT, liu2021groupfree, wang2022detr3d, zhao2021point_transformer}. Since 3D data and annotation are scarcer and more expensive than the 2D counterparts, which makes it hard to train plain transformers, previous works inject inductive bias by using, \eg, hierarchical sub-sampling and local attention. 

Although the modified transformers have achieved impressive results, a strong baseline, which shows the potential of plain transformers in point cloud understanding, is still missing. 
We believe that plain transformers are worth more research interest in the context of 3D computer vision. 
First, a plain transformer consists of fundamental operations, \eg, matrix multiplication and the softmax activation, which makes it relatively easy to deploy on hardware~\cite{zhong:2023:transformer, qi:2021:accelerating}. However, modified transformers are non-trivial to deploy due to special operations~\cite{lin:2021:pointacc}, \eg, the modified attention.
Furthermore, the local attention provides inductive bias but also hurts the capability to capture long-range dependencies, which are important in some tasks, \eg, detecting large objects from point clouds~\cite{votenet_Qi_2019_ICCV, liu2021groupfree}. 
Moreover, multi-modal transformers have invoked research interest recently, as they unify language, vision, and audio understanding~\cite{baevski2022data2vec, kim2021vilt, ramesh2021dall_e, reed2022generalist}. 
Although the inductive bias improves performance on one specific modality, it usually cannot generalize to others~\cite{baevski2022data2vec}. Thus, a baseline of plain transformers for point clouds is necessary for future research on multi-modal models. 

Another issue of previous works is the complexity of evaluation tasks. Many works~\cite{engel2021point_transformer, pos_bert, guo2021pct, point_mae, point_bert, zhao2021point_transformer} focus on classification using either clean synthetic data, \eg, ShapeNet dataset~\cite{shapenet_dataset} or single-object real-world data, \eg, ScanObjectNN dataset~\cite{uy-scanobjectnn-iccv19}. We speculate that these data and tasks are too simple to convincingly justify the network design and show the full potential of transformers, which are known to have a large model capacity. 
Also, the design based on simple data might not generalize well on complex real-world point clouds, which limits the application of transformers in real-world tasks, \eg, robotics and autonomous driving. 
Moreover, due to the quadratic complexity of multi-head attention~\cite{vaswani2017transformer}, plain transformers are usually computationally expensive for real-world 3D data. However, this problem could be neglected if solely small point clouds are studied. 

In this work, we revisit the design of plain transformers and evaluate our methods on complicated large-scale real-world point clouds. To narrow the scope of this work, we focus on transformers as backbones and do not consider the usage as task-specific necks or heads~\cite{liu2021groupfree, wang2022detr3d}. 
While keeping the overall architecture plain, we optimize some components of transformers for point clouds, \eg, the patchifier and position embedding. 
We systematically investigate existing patchifiers, \eg, ball query and kNN.
Also, we introduce Farthest Point Clustering (FPC) to study the effect of non-overlapping patchifers. 
Furthermore, we propose incorporating global information into position embedding to better describe the patches' position. 
To close the performance gap caused by lack of inductive bias, we also explore the self-supervised pre-training of our models. 
Based on the successful masked autoencoder (MAE)~\cite{mae}, we propose a novel method \emph{drop patch}. It suppresses the information leakage caused by the position embedding in the decoder by only reconstructing a proportion of unseen patches. This simple method significantly improves the results of pre-training and reduces the computation. 

The contribution of our work is many-fold:
\begin{enumerate}
    \item We optimize some essential components of plain transformers, \eg, the patchifier and position embedding, for more effective point cloud understanding. 
    \item We investigate the information leakage problem in the standard masked autoencoder for 3D vision and propose drop patch for better transfer learning results. 
    \item We focus on complex real-world point clouds to evaluate our designs.
    \item We show that with proper designs and self-supervised pre-training, plain transformers can achieve SOTA results in real-world 3D object detection and semantic segmentation while being efficient. 
\end{enumerate}

\section{Related Works}
\noindent
\textbf{Transformers for Point Clouds.}
Many previous works modify the architecture of vision transformer (ViT)~\cite{dosovitskiy2020vit} for point cloud understanding. Common approaches are applying local attention and down-sampling. For instance, \cite{engel2021point_transformer, pan2021pointformer, park2022fast, zhao2021point_transformer} limit the attention mechanism in a local region, which integrates the locality into transformers and reduces the computational cost. Also, Hui \etal~\cite{Hui2021pyrimidPCT} perform hierarchical down-sampling to build a pyramid architecture for large-scale point clouds. 
PatchFormer~\cite{Zhang2022patchformer} down-samples the queries to improve efficiency. PCT~\cite{guo2021pct} uses transformers to aggregate high-level features after set abstraction modules~\cite{pointnet++}.
On the contrary, we intend to keep the transformer plain in this work. We use multi-head attention~\cite{vaswani2017transformer} globally and only down-sample point clouds once for patchifying. 

\noindent
\textbf{Pre-Training without 3D Annotation.}
A lot of works have investigated the pre-training without 3D annotation to improve convergence and performance in 3D vision tasks. 
Some works attempt to directly initialize 3D networks using pre-trained 2D models, \eg, by mapping weights of 2D ConvNets to 3D ones~\cite{image2point} or adopting a pre-trained ViT~\cite{pix4point}. Also, PointCLIP~\cite{zhang2022pointclip} utilizes pre-trained CLIP-models~\cite{radford2021clip} to classify point clouds. 
Contrastive methods are usually based on the invariance of 3D features. Previous works use invariances to create a correspondence between two point clouds viewed from different view angles~\cite{PointContrast_eccv_2020, exploring_efficient}, between point clouds and color images~\cite{learning_from_2d}, between voxels and point clouds~\cite{depthcontrast} or between depth maps and point clouds~\cite{li2022closer}. Also, 4DContrast~\cite{4dcontrast} uses dynamic spatial-temporal correspondence in pre-training. 
Generative methods restore missing information from partially visible inputs. Wang \etal~\cite{wang2021occlusion} reconstruct complete point clouds from occluded single-view ones. Point-BERT~\cite{point_bert} follows the successful BERT~\cite{devlin2018bert} framework to predict the missing tokens from masked point clouds. POS-BERT~\cite{pos_bert} combines the BERT pipeline with momentum tokenizers and contrastive learning. Following masked autoencoder (MAE)~\cite{mae}, Point-MAE~\cite{point_mae} reconstructs the coordinates of masked points. Point-M2AE~\cite{point_m2ae} extends the MAE pipeline to hierarchical multi-scale networks. MaskPoint~\cite{masked_discrimi} models an implicit representation to avoid information leakage.

\section{Methods}
In this section, we first review the basic architecture of plain transformers for point clouds (Section~\ref{subsec:plain_tr}). Then, we investigate two crucial but long-overlooked components in plain transformers, \ie, the patchifier (Section~\ref{subsec:patch}) and position embedding (Section~\ref{subsec:pos_embd}). Later, we show how to pre-train our models using self-supervision (Section~\ref{subsec:ssp}).

\subsection{Plain Transformers for Point Clouds}
\label{subsec:plain_tr}

As shown in Figure~\ref{fig:transformer}, a plain transformer can be separated into five components: a patchifier, patch embedding, position embedding, a transformer encoder consisting of multiple transformer layers, and a task-specific head. The patchifier divides the input point cloud into small patches. 
The patch embedding encodes each point patch into a feature vector. A PointNet~\cite{fpointnet_Qi_2018_CVPR} is usually used for patch embedding~\cite{masked_discrimi, misra2021_3detr, point_mae, point_bert}. All patch features build up a sequence, which is then fed into the transformer encoder. Since the multi-head attention is permutation-equivariant and unaware of the position of each patch, transformers require position embedding~\cite{vaswani2017transformer}, which directly injects positional information into the sequence. The transformer encoder then extracts informative features, which are utilized by the task-specific head.

\subsection{Patchifier}
\label{subsec:patch}
The process to build patches (\ie patchify) can be further separated into \emph{sampling} and \emph{grouping}. Without loss of generality, we only consider inputs with 3D coordinates and ignore other channels, \eg, colors,  because they do not affect patchifying and are assigned to respective coordinates afterward~\cite{masked_discrimi, misra2021_3detr, point_mae, pointnet++, point_bert}. Given a point cloud $\{ x_i | x_i \in \mathbb{R}^{3} \}_{i=1}^N$ with $N$ points, the patchifier first sub-samples $M$ key points $\{ s_i | s_i \in \mathbb{R}^{3} \}_{i=1}^M$ using farthest point sampling (FPS)~\cite{pointnet++}. Then, the patchifier searches $K$ neighbors for each key point to build patches $\{ \mathbf{P}_i \}_{i=1}^{M}$ with $| \mathbf{P}_i | = K$. In previous works, ball query~\cite{pointnet++, misra2021_3detr} and k-Nearest-Neighbor (kNN)~\cite{point_bert, point_mae, pix4point, masked_discrimi} are used for grouping. The former searches $K$ points in a sphere with a given radius around each key point, while the latter assigns $K$ closest neighbors to each key point. Then, each patch $\mathbf{P}_i$ is encoded into a $C$-dimensional feature vector $f_i \in \mathbb{R}^C$ by the patch embedding, which is usually a shared PointNet. 

Despite the different choices of patchifiers, previous works usually use a large patch number $M$ with $N \ll MK$. For instance, 3DETR~\cite{misra2021_3detr} divides an input of 40K points into 2048 patches, which is an order of magnitude greater than a common ViT~\cite{dosovitskiy2020vit}. As the complexity of the multi-head attention is quadratic to the sequence length, it results in high computational costs, which limits the application of plain transformers in point cloud understanding, especially for large real-world point clouds. Also, the patchifiers in previous works generate overlapping patches. Although such a design can improve the stability of plain transformers~\cite{xiao2021early}, it causes information leakage during pre-training with MAE, since the masked and reserved patches might share points~(see Section~\ref{subsec:ssp}). 

To our best knowledge, the impact of shorter sequences and different choices of patchifiers have not drawn much attention in previous research. In our work, we use a shorter sequence with $N \approx MK $ to improve the efficiency of plain transformers. Also, we systematically compare different patchifiers with various setups. In addition to the aforementioned two overlapping patchifiers, we evaluate non-overlapping ones, \eg, k-means and our proposed method Farthest Point Clustering (FPC). 

\noindent
\textbf{Farthest Point Clustering. } We still use FPS to sample $M$ key points $\{ s_i\}^M_{i=1}$. 
We cluster the $N$ input points into $M$ patches by assigning each point $x_i$ to its nearest key point $s_i$. 
Note that, unlike kNN, each point is assigned to only one key point, so that the generated patches do not overlap. 
Then, we further sample $K$ points in each cluster so that each patch has the same number of points, following ball query~\cite{pointnet++}. This algorithm's pseudo-code and implementation details are provided in the supplementary material. 

\begin{figure}[t]
    \centering
    \includegraphics[width=0.47\textwidth]{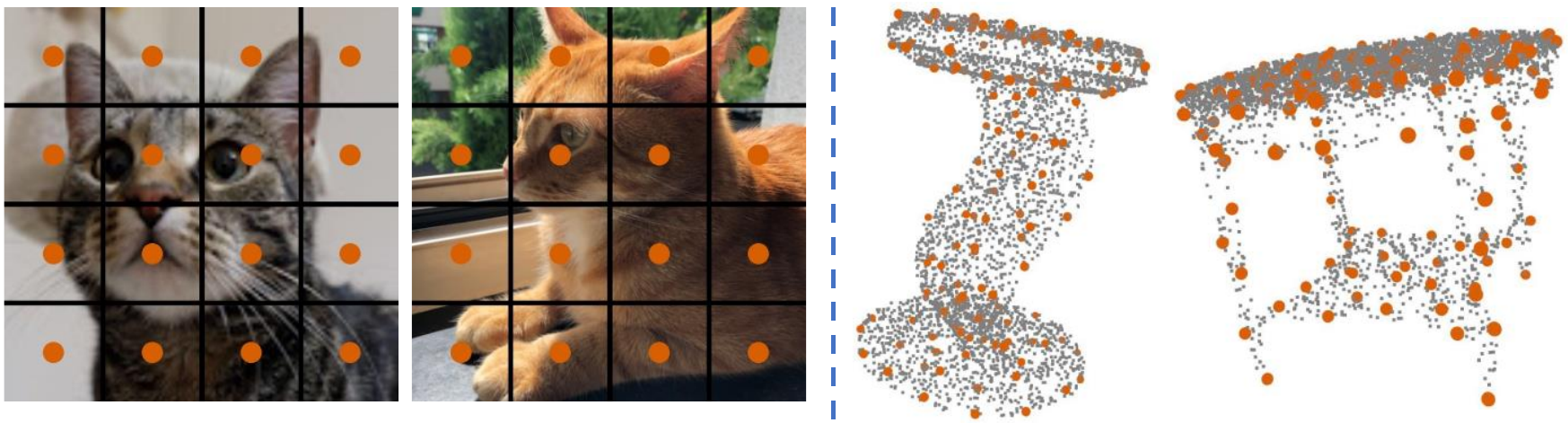}
    \caption{``Positions'' of patches (orange dots) in different data. In images, they are independent of the content. The ``positions'' alone contain almost no information. In point clouds, ``positions'' are unique for each data sample and thus more informative, \ie, one can know how the point cloud roughly looks like by only observing the ``positions''.}
    \label{fig:pos}
\end{figure}

\subsection{Position Embedding}
\label{subsec:pos_embd}
Position embedding is a mapping $\mathbb{R}^3 \to \mathbb{R}^C $, which encodes the coordinate of each key point into a vector: 
\begin{align}
    \label{eq:pos_embed}
    e_i = \mathrm{PosEmbed}(s_i)
\end{align}
\noindent 
Previous works use Fourier features \cite{misra2021_3detr, tancik2020fourier} or multi-layer perceptron (MLP) \cite{point_mae, masked_discrimi} as position embedding for point clouds. 
They all treat each position $s_i$ separately, as formulated in Equation~\ref{eq:pos_embed}, and neglect the global information in all key points $\{s_i\}_{i=1}^M$. While the ``positions'' in natural languages and images are fixed and shared across all data samples, they are content-dependent and more informative in point clouds, as shown in Figure~\ref{fig:pos}. 
Our intuition is that the global information in position embedding might benefit point cloud understanding since it directly makes each patch aware of others' positions.
In this work, we first transform each coordinate $s_i$ into a high dimensional space using an MLP. Then we aggregate the global feature via global max pooling. The global feature is then concatenated to each coordinate and further projected with another MLP. Our position embedding can be formulated as follows: 
\begin{figure*}[t]
    \centering
    \begin{tabular}{ccc}
         \includegraphics[width=0.3\textwidth]{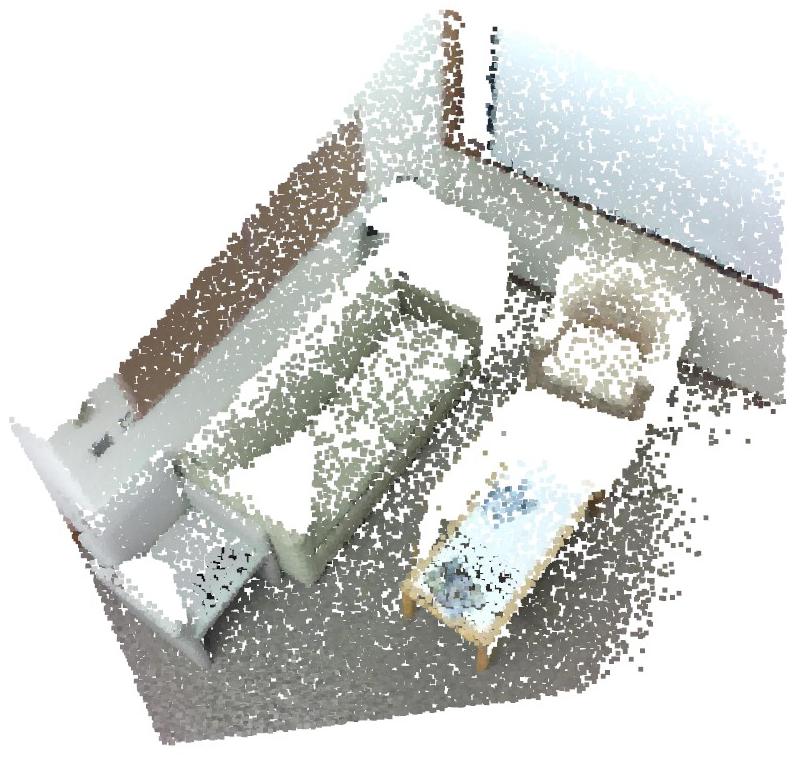}
         &
         \includegraphics[width=0.3\textwidth]{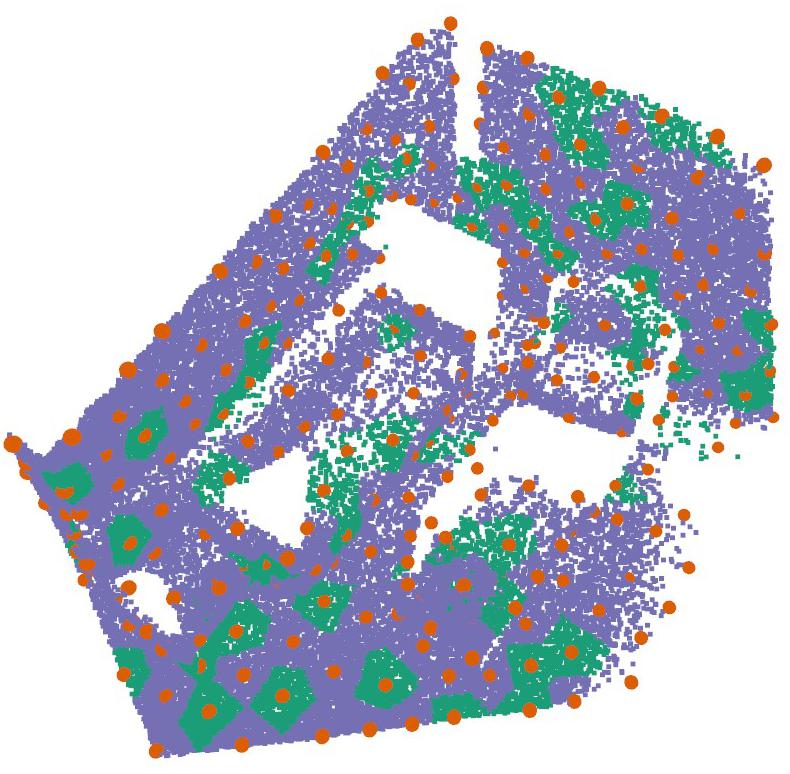}
         &
         \includegraphics[width=0.3\textwidth]{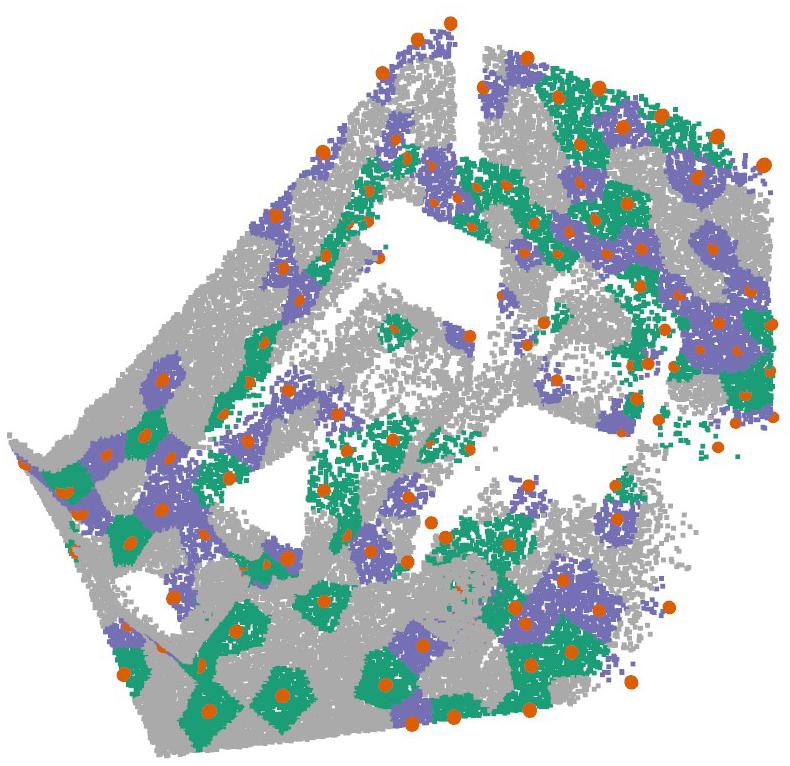} \\
         (a) Original point cloud & (b) Standard MAE & (c) MAE with drop patch
    \end{tabular}
    \caption{Illustration of drop patch for point cloud MAE. 
    \textcolor{pgreen}{\textbf{Green patches}} are reserved and fed into the encoder. 
    \textcolor{ppurple}{\textbf{Purple patches}} are masked out and to be reconstructed. 
    \textcolor{pgrey}{\textbf{Grey patches}} are dropped and neglected by both the encoder and decoder. 
    \textcolor{porange}{\textbf{Orange dots}} are key points visible for the MAE decoder (\ie, embedded in the query). 
    In the standard MAE, the \textcolor{porange}{\textbf{key points}} leak much positional information of \textcolor{ppurple}{\textbf{masked patches}} to the decoder since the overall shape of the complete point cloud can be recognized by only observing these points. The MAE decoder can up-sample the \textcolor{porange}{\textbf{key points}} to accomplish the reconstruction task in MAE pre-training, instead of extracting informative features from the \textcolor{pgreen}{\textbf{reserved patches}}. With drop patch, the decoder sees much fewer \textcolor{porange}{\textbf{key points}} and is forced to learn patch features.}
    \label{fig:drop_path}
\end{figure*}
\begin{align}
    g_i &= \mathrm{MLP}_1 (s_i) \\
    g &= \mathrm{MaxPool}(g_1, ..., g_i, ..., g_M) \\
    e_i &= \mathrm{MLP}_2 (\mathrm{Concat}(g, s_i))
\end{align}
Then, $e_i$ is added to its respective patch feature $f_i$, following the common practice in previous works. Note that in pre-training with MAE (Section~\ref{subsec:ssp}), the global pooling in the encoder aggregates global features $g$ only from visible patches. Thus, the pooling operation does not leak information about masked patches in pre-training.

\subsection{Self-Supervised Pre-Training}
In this work, we use MAE to pre-train our models. 

\label{subsec:ssp}
\noindent
\textbf{Masked Autoencoders for Point Clouds.} The idea of MAE~\cite{mae} is to randomly divide input patches $\{ \mathbf{P}_i \}_{i=1}^M$ into two disjoint subsets $\{\mathbf{R}_i\}$ and $\{\mathbf{M}_i\}$. Patches $\{\mathbf{M}_i\}$ are masked out, and the transformer encoder only sees the reserved patches $\{\mathbf{R}_i\}$. 
With a transformer-based decoder, the model is trained to reconstruct the masked patches $\{\mathbf{M}_i\}$ based on their positions and visible patches. 
After pre-training, the decoder is abandoned, and the encoder (with patch embedding, position embedding, \etc) can be used for downstream tasks. He~\etal~\cite{mae} propose using a large mask ratio (\eg, \SI{75}{\percent}) for good performance. 

However, for point clouds, MAE encounters two possible information leakage problems. On the one hand, the patches might overlap with each other, \ie, $\{\mathbf{R}_i\}$ might share points with $\{\mathbf{M}_i\}$, making the pre-training less effective. MaskPoint~\cite{masked_discrimi} suggests using an extremely high mask ratio (\eg, \SI{90}{\percent}) as a workaround. 
With non-overlapping patchifiers, \eg, k-means and FPC, this problem can be avoided. 
On the other hand, the decoder uses the positional information of both masked and reserved patches as the query. 
As discussed in Section~\ref{subsec:pos_embd}, the position embedding of point clouds corresponds to the sub-sampled input~(\ie, key points) and leaks the positional information of the points to be reconstructed. In this case, reconstructing the masked patches is equivalent to up-sampling the key points and becomes trivial (see Figure~\ref{fig:drop_path}).

\noindent
\textbf{Drop Patch.} 
To address the information leakage in the decoder, Liu~\etal~\cite{masked_discrimi} discriminate if a randomly generated point is close enough to the original input point cloud, instead of reconstructing masked patches directly. However, this method is still complex and has more hyper-parameters, \eg, the distance threshold and distribution of the random points.
On the contrary, we propose an awkwardly simple yet effective method. 
For each iteration, we randomly split input patches $\{ \mathbf{P}_i \}_{i=1}^M$ into three disjoint sets $\{\mathbf{D}_i\}$, $\{\mathbf{R}_i\}$ and $\{\mathbf{M}_i\}$, instead of two. Then, patches $\{\mathbf{D}_i\}$ are immediately dropped. The transformer decoder reconstructs $\{\mathbf{M}_i\}$ by using features from $\{\mathbf{R}_i\}$ and the positional information of both $\{\mathbf{M}_i\}$ and $\{\mathbf{R}_i\}$. We name this method \emph{drop patch}.
With enough patches dropped, the decoder sees too few key points to perform the trivial up-sampling. 
In this work, we use $|\{\mathbf{D}_i\}|:|\{\mathbf{R}_i\}|:|\{\mathbf{M}_i\}|=2:1:1$, which is similar to the original MAE with a mask ratio of \SI{75}{\percent}, as the encoder sees \SI{25}{\percent} patches in both cases. The principle of drop patch is illustrated in Figure~\ref{fig:drop_path}. Note that drop patch also reduces the patches to be reconstructed and thus decreases the computational cost of pre-training.

\noindent
\textbf{Loss Function.}  After the decoder, we use a fully connected layer to generate a prediction. For each masked patch consisting of a key point and its $K$ neighbors, we predict $K$ offsets from the key point to its neighbors. 
We apply L2 Chamfer distance as loss function and only apply it on masked patches, following~\cite{mae}. 


\section{Experiments}
We first introduce our experiment setups in Section~\ref{subsec:setups}. Then, we show our main results compared with SOTA in Section~\ref{subsec:main_results}. After that, we justify our design choices of patchifiers, position embedding, and drop patch with extensive ablation studies in Section~\ref{subsec:ablation}. Also, we compare the efficiency of our models with previous works. 

\subsection{Setups}
\label{subsec:setups}
In this work, we use a transformer encoder with 3 layers as the backbone if it is not otherwise specified. Each transformer layer has 256 channels and 4 heads, while the feed-forward sub-net has 512 channels. Unlike ViT, we don not use the class token~\cite{dosovitskiy2020vit}. For all experiments, we use an AdamW optimizer~\cite{loshchilov2017decoupled} with a weight decay of 0.01, the cosine annealing schedule~\cite{loshchilov2016sgdr} and gradient clip of 0.1. All training is warmed up for 10 epochs. Other task-specific configurations are explained as follows. More technical details are provided in our supplementary material. 

\noindent
\textbf{Pre-Training.} We use a decoder with 2 transformer layers. Each layer has 256 channels and 4 heads. The feed-forward dimension is 256. We use ScanNet~\cite{dai2017scannet} to pre-train our models. The dataset consists of $\sim$2.5M frames of RGB-D images captured in 1513 indoor scenes. We sample every 25 frames from the train set, following previous works~\cite{exploring_efficient, li2022closer, PointContrast_eccv_2020}. For each frame, we randomly sample 20K points for pre-training. Our patchifier divides each point cloud into 256 patches and samples 128 points in each patch. 
We use an initial learning rate of $5\times10^{-4}$ and train for 120 epochs with a batch size of 64. 
The color channels are handled differently in the pre-training since a lot of object detectors~\cite{HGNet_Chen_2020_CVPR, misra2021_3detr, votenet_Qi_2019_ICCV,  PointContrast_eccv_2020, depthcontrast} do not use color information, whereas the models for semantic segmentation methods do~\cite{mink_Choy_2019_CVPR, pointnet++, qian2022pointnext, kpconv_thomas_2019_iccv, PointContrast_eccv_2020, depthcontrast}. 
For object detection, we only use 3D coordinates (\ie, geometry) in pre-training. For semantic segmentation, we pre-train with geometry and color. However, we do not reconstruct color channels, as we empirically find it has no significant effect. 

\noindent
\textbf{Object Detection.} We adopt the detection pipeline from 3DETR~\cite{misra2021_3detr}, an end-to-end transformer-only detector consisting of 3 encoder layers and 8 decoder layers. We simply replace the encoder with our plain transformers. Other configurations are as same as 3DETR. We train detectors on ScanNet~\cite{dai2017scannet}. We follow the official train/val split and use 1201 multi-view point clouds for training and 312 for validation. 
As input, we randomly sample 40K points. Point clouds are divided into 512 patches with 128 points ($M=512$ and $K=128$). 
All models are trained for 1080 epochs with an initial learning rate of $5\times10^{-4}$ and a batch size of 8. 
Metrics are mean average precision with 25\%- and 50\% 3D-IoU threshold (\ie, AP25 and AP50) over 18 representative classes. 

\noindent
\textbf{Semantic Segmentation.}
Since the segmentation task requires point-wise output, we up-sample the features from the transformer encoder using nearest neighbor interpolation~\cite{pointnet++}. The point-wise features are further projected by a shared MLP and fed into an MLP-based prediction head. We evaluate our models on the S3DIS~\cite{s3dis_dataset_17} dataset, which consists of real-world scans from 6 large indoor areas. Following previous works, we report the validation results on Area 5 and train models in other areas. 
Due to the large size of each point cloud, we voxelize the point clouds with a voxel size of \SI{4}{\cm} and randomly crop 24K points for each forward pass. We use $M=512$ and $K=64$. We apply the same data augmentation as~\cite{qian2022pointnext}. All models are trained for 300 epochs with a batch size of 16. Metrics are mean accuracy~(mAcc) and mean IoU (mIoU) over 13 classes. 

\begin{table}[t]
    \centering
    \begin{tabular}{lcccc}
    \toprule
        \textbf{Methods} & \textbf{Pre.} & \textbf{Tr.} & \textbf{AP25} & \textbf{AP50} \\
    \midrule
        VoteNet~\cite{votenet_Qi_2019_ICCV} & & & 58.6 & 33.5 \\
        PointContrast~\cite{PointContrast_eccv_2020} & \cmark & & 59.2 & 38.0 \\
        Hou~\etal~\cite{exploring_efficient} & \cmark & & - & 39.3 \\
        4DContrast~\cite{4dcontrast} & \cmark & & - & 38.2 \\ 
        DepthContrast ($\times$1)~\cite{depthcontrast} & \cmark & & 61.3 & - \\
        DepthContrast ($\times$3)~\cite{depthcontrast} & \cmark & & 64.0 & 42.9 \\
        DPCo~\cite{li2022closer} & \cmark & & 64.2 & 41.5 \\
    \midrule
        3DETR~\cite{misra2021_3detr} & & \cmark & 62.1 & 37.9 \\
        PointFormer~\cite{pan2021pointformer} & & \cmark* & 64.1 & 42.6 \\
        MaskPoint (L3)~\cite{masked_discrimi} & \cmark & \cmark & 63.4 & 40.6 \\
        MaskPoint (L12)~\cite{masked_discrimi} & \cmark & \cmark & 64.2 & 42.1 \\
        \rowcolor{gray!25}
        \textbf{Ours (512 patches)} & & & & \\
        \textit{-- from scratch}& & \cmark & 61.6 & 38.8 \\
        \textit{-- MAE}& \cmark & \cmark & 62.7 & 42.2 \\
        \textit{-- MAE + DP}& \cmark & \cmark & 64.1 & 43.0 \\

        \rowcolor{gray!25}
        \textbf{Ours (1024 patches)} & & & & \\
        \textit{-- from scratch}& & \cmark & 62.4 & 41.3 \\
        \textit{-- MAE}& \cmark & \cmark & 64.6 & 44.8 \\
        \textit{-- MAE + DP}& \cmark & \cmark & \textbf{65.6} & \textbf{45.3} \\

    \bottomrule
    \end{tabular}
    \caption{Object detection results on ScanNet V2 validation set. AP25 und AP50 are in percentage. Pre.: pre-trained. Tr.: transformer-based. DP: drop patch. Mark \cmark*: with local attention. }
    \label{tab:det}
\end{table}

\subsection{Results and Analysis}
\label{subsec:main_results}
\noindent
\textbf{Object Detection.}
We first compare our results with SOTA methods in object detection on ScanNet (Table~\ref{tab:det}). 
With 512 patches, our detector without pre-training performs similarly to the original 3DETR with 2048 patches. 
This demonstrates that it is possible to use a shorter sequence length without a significant performance drop. 
With MAE, our results are improved by \SI{1.1}{\percent} AP25 and \SI{3.4}{\percent} AP50~(absolute), showing the power of pre-training. Drop patch further raises the AP25 by \SI{1.4}{\percent} and AP50 by \SI{0.8}{\percent}. 
Our results with 512 patches (\SI{64.1}{\percent} AP25 and \SI{43.0}{\percent} AP50) surpass the previous SOTA MaskPoint (L3 variant, \ie, with 3 encoder layers) with a clear margin while showing similar performance as the heavy 12-layer variant. 
We further evaluate our models using 1024 patches. The model already surpasses 3DETR without pre-training. When pre-trained, it achieves \SI{65.6}{\percent} AP25 and \SI{45.3}{\percent} AP50, significantly outperforming previous works. Note that we use FPC for 512 patches, but ball query for 1024 patches since FPC brings sub-optimal results with a longer sequence. More discussions are provided in Section~\ref{subsec:ablation}. 

\begin{table}[t]
    \centering
    \begin{tabular}{lcccc}
        \toprule
        \textbf{Methods} & \textbf{Pre.} & \textbf{Tr.} & \textbf{mAcc} & \textbf{mIoU} \\
        \midrule
        PointNet++~\cite{pointnet++} & & & - & 53.5 \\
        MinkowskiNet-32~\cite{mink_Choy_2019_CVPR} & & & 71.7 & 65.4 \\
        KPConv~\cite{kpconv_thomas_2019_iccv} & & & 72.8 & 67.1 \\
        PointNeXt-B~\cite{qian2022pointnext} & & & 74.3 & 67.5 \\
        PointNeXt-L~\cite{qian2022pointnext} & & & 76.1 & 69.5 \\
        pixel-to-point~\cite{learning_from_2d} & \cmark &  & 75.2 & 68.3 \\
        PointContrast~\cite{PointContrast_eccv_2020} & \cmark & & - & 70.3 \\
        DepthContrast~\cite{depthcontrast} & \cmark & & - & \textbf{70.9} \\

        \midrule
        PCT~\cite{guo2021pct} & & \cmark* & 67.7 & 61.3 \\
        PatchFormer~\cite{Zhang2022patchformer} & & \cmark* & - & 68.1 \\
        PointTransformer~\cite{zhao2021point_transformer} & & \cmark* & 76.5 & 70.4 \\
        Pix4Point~\cite{pix4point} & \cmark & \cmark & 73.7 & 67.5 \\
        
        \rowcolor{gray!25}
        \textbf{Ours (3 layers)} & & & & \\
        \textit{-- from scratch}& & \cmark & 66.4 & 60.0 \\
        \textit{-- MAE}& \cmark & \cmark & 73.6 & 67.2 \\
        \textit{-- MAE + DP}& \cmark & \cmark & 74.7 & 67.6 \\
        \rowcolor{gray!25}
        \textbf{Ours (12 layers)} & & & & \\
        \textit{-- from scratch}&  & \cmark & 70.0 & 63.2 \\
        \textit{-- MAE}& \cmark & \cmark & 75.9 & 69.5 \\ 
        \textit{-- MAE + DP}& \cmark & \cmark & \textbf{77.0} & 70.4 \\      
        \bottomrule
    \end{tabular}
    \caption{Semantic segmentation on S3DIS dataset Area 5. Reported mAcc and mIoU are in percentage. DP: drop patch. Mark \cmark*: with modified transformers. Our models use 512 patches. }
    \label{tab:seg}
\end{table}

\noindent
\textbf{Semantic Segmentation.}
We report the semantic segmentation results in Table~\ref{tab:seg}. While the performance without pre-training is low~(\SI{66.4}{\percent} mAcc and \SI{60.0}{\percent} mIoU), the standard MAE improves the metrics by \SI{7.2}{\percent} and \SI{7.2}{\percent}, respectively. Since the S3DIS dataset is relatively small, we believe the results on this dataset benefit more from the pre-training. Also, drop patch further improves the mAcc und mIoU by \SI{1.1}{\percent} and \SI{0.4}{\percent}, respectively. 
When scaled up from 3 to 12 layers, our model achieves significantly better results with \SI{77.0}{\percent} mAcc and \SI{70.4}{\percent} mIoU. The performance surpasses some highly optimized models, \eg, PointTransformer~\cite{zhao2021point_transformer} and PointNeXt~\cite{qian2022pointnext}. It implies that self-supervised pre-training brings comparable improvement to architecture optimization. 

\subsection{Ablation Studies and Computational Costs}
\label{subsec:ablation}
We conduct ablations studies primarily on the object detection task, as object detection with plain transformers is better understood in previous works~\cite{masked_discrimi, misra2021_3detr}. Also, we use AP25 as the primary metric, following~\cite{misra2021_3detr}.

\noindent
\textbf{Patchifiers.} With this ablation study, we attempt to clarify the impact of different patchifiers. Their interaction with position embedding, pre-training, and patch numbers is also researched. 
\begin{table}[t]
    \centering
    \begin{tabular}{ccccccc}
    \toprule
       \textbf{ID} & \textbf{Group} & $M$ & \textbf{Pre} & \textbf{PE} & \textbf{AP25} & \textbf{AP50} \\
    \midrule
       1 & Ball & 512 & & & 59.8 & 37.9 \\
       2 & kNN & 512 & & & 60.8 & 38.0 \\
       3 & k-means & 512 & & & 59.5 & 36.3 \\
       4 & FPC & 512 & & & 60.3 & 38.1 \\
    \midrule
       5 & Ball & 512 & & \cmark & 61.1 & 39.7 \\
       6 & kNN & 512 & & \cmark & 61.7 & 41.0 \\
       7 & k-means & 512 & & \cmark & 60.2 & 34.0 \\
       8 & FPC & 512 & & \cmark & 61.6 & 38.8 \\
    \midrule
       9 & Ball & 512 & \cmark & \cmark & 63.4 & 42.1 \\
       10 & kNN & 512 & \cmark & \cmark & 63.7 & 42.4 \\
       11 & k-means & 512 & \cmark & \cmark & 62.7 & 38.7 \\
       12 & FPC & 512 & \cmark & \cmark & \textbf{64.1} & \textbf{43.0} \\
    \midrule
       13 & Ball & 1024 & & \cmark & 62.4 & 41.3 \\
       14 & kNN & 1024 & & \cmark & 63.5 & 39.9 \\
       15 & k-means & 1024 & & \cmark & 59.0 & 36.6 \\
       16 & FPC & 1024 & & \cmark & 61.6 & 36.9 \\
    \midrule
       17 & Ball & 1024 & \cmark & \cmark & \textbf{65.6} & \textbf{45.3} \\
       18 & kNN & 1024 & \cmark & \cmark & 65.0 & 43.5 \\
       19 & k-means & 1024 & \cmark & \cmark & 63.8 & 40.3 \\
       20 & FPC & 1024 & \cmark & \cmark & 64.6 & 44.3 \\
    \bottomrule
    \end{tabular}
    \caption{Ablation study on patchifiers. Drop patch is applied for pre-training. Global information is used in position embedding. Group: grouping methods. $M$: number of patches. Pre: pre-trained or not. PE: with position embedding or not. \textbf{Bold}: best results with different patch numbers. }
    \label{tab:patchifier}
\end{table}
As shown in Table~\ref{tab:patchifier}, k-means achieves the worst performance with all setups. It is because k-means is sensitive to the spatial density of points. 
Since real-world point clouds are usually captured with depth sensors and the point density varies with depth, k-means leads to irregular patch sizes and is sub-optimal. 

When models are not pre-trained, the kNN patchifier achieves the best performance (experiment 2, 6, and 14). Similar results are also observed in image processing, where early convolutions improve the performance of a standard ViT~\cite{xiao2021early}. However, when models are pre-trained with MAE, it is sub-optimal compared to FPC (experiment 10 and 12). Since kNN generates overlapping patches, it leaks the information of points to be reconstructed and thus degrades the effect of MAE. 

FPC performs best when the patch number is small~(\eg, 512) and models are pre-trained. However, when it comes to 1024 patches, it is inferior compared to kNN and ball query. Since patches cannot overlap, FPC generates small and irregular patches in this case, which harms the performance. 

Ball query outperforms other methods for a large patch number (\eg, 1024), because it guarantees a consistent scale and shape of patches, which helps models learn spatial features. 
Such an advantage is also reported in~\cite{kpconv_thomas_2019_iccv}. However, ball query is sub-optimal for a small patch number (\eg, 512) since it is difficult to set a suitable radius in this case. While the patch embedding cannot capture fine-grained details with a large radius, the patches cannot cover the entire point clouds with a small radius. 

To sum up, the performance of patchifiers is often affected by competing factors, which makes the optimal option of patchifiers conditional. 
Depending on patch numbers and pre-training, ball query, FPC, and kNN can deliver the best result. 
We pay more attention to pre-trained models, as pre-training is crucial to compensate for the performance gap due to the lack of inductive bias. 
Thus, this work uses FPC for a smaller patch number ($M\leq512$) and ball query for a larger patch number~($M>512$).

\begin{table}[t]
    \centering
    \setlength\tabcolsep{4.2pt}
    \begin{tabular}{cccccccc}
    \toprule
       \textbf{ID} & \textbf{Group} & $M$ & \textbf{Pre} & \textbf{PE} & \textbf{Add} &  \textbf{AP25} & \textbf{AP50} \\
    \midrule
       1 & Ball & 512 & & - & - & 59.8 & 37.9 \\
       2 & Ball & 512 & \cmark & - & - & 60.4 & 38.3 \\
       3 & FPC & 512 & & - & - & 60.3 & 38.1 \\
       4 & FPC & 512 & \cmark & - & - & 59.7 & 37.2 \\
    \midrule
       5 & FPC & 512 & & Fourier & first & 59.9 & 38.6 \\
       6 & FPC & 512 & & MLP & first & 61.1 & 37.9 \\
       7 & FPC & 512 & & Global & first & 61.6 & 38.8 \\
       8 & FPC & 512 & \cmark & Fourier & first & 61.6 & 40.9 \\
       9 & FPC & 512 & \cmark & MLP & first & 62.4 & 42.6 \\
       10 & FPC & 512 & \cmark & Global & first & \textbf{64.1} & \textbf{43.0} \\
    \midrule
       11 & FPC & 512 & & Fourier & all & 60.3 & 38.6 \\
       12 & FPC & 512 & & MLP & all & 60.7 & 39.0 \\
       13 & FPC & 512 & & Global & all & 61.3 & 36.7 \\
       14 & FPC & 512 & \cmark & Fourier & all & 61.4 & 39.2 \\
       15 & FPC & 512 & \cmark & MLP & all & 61.4 & 38.6 \\
       16 & FPC & 512 & \cmark & Global & all & 63.3 & 42.0 \\
    \midrule
       17 & Ball & 1024 & & MLP & first & 62.1 & 40.1 \\
       18 & Ball & 1024 & & Global & first & 62.4 & 41.3 \\
       19 & Ball & 1024 & \cmark & MLP & first & 64.3 & 44.0 \\
       20 & Ball & 1024 & \cmark & Global & first & \textbf{65.6} & \textbf{45.3} \\
    \bottomrule
    \end{tabular}
    \caption{Ablation study on position embedding. Drop patch is applied in pre-training. Group: grouping methods of patchifiers. $M$: number of patches. Pre: pre-trained or not. PE: type of position embedding. Add: the encoder layers where the position embedding is added. \textbf{Bold}: best results with different patch numbers.}
    \label{tab:pos_embed}
\end{table}

\noindent
\textbf{Position Embedding.} With this ablation study, we systematically compare different types of position embedding. Besides Fourier features, MLP, and our method with global information, we also evaluate models without position embedding in the transformer encoder. Note that besides the transformer encoder (\ie, the backbone), the decoder in MAE and the detection head in 3DETR also require position embedding. For simplicity, we use the same type of position embedding in the transformer encoder, the MAE decoder, and the detection head. For variants without position embedding in the encoder, we use Fourier features for other components, following~\cite{misra2021_3detr}. We primarily use FPC in this ablation study to highlight the impact of position embedding since overlapping patchifiers can implicitly encode the relative position of patches~\cite{misra2021_3detr}. 

Comparing experiment 3, 5, 6, and 7 in Table~\ref{tab:pos_embed}, one can see that Fourier features degrade the performance when trained from scratch, which is also observed in previous work~\cite{misra2021_3detr}. On the contrary, MLP and our method bring significant improvement compared to the variant without position embedding. Also, experiment 1-4 show that pre-training is ineffective if position embedding is not added. It is feasible since the positional information of input patches is necessary for the reconstruction task in MAE. On the other hand, experiment 8-10 show that position embedding makes the pre-training more effective. Meanwhile, the results in 5-10 show that parametric position embedding (\ie, MLP and Global) performs better than the non-parametric Fourier features. Also, our method performs better than MLP, which verifies our intuition in Section~\ref{subsec:pos_embd} that the global information in position embedding is beneficial. The results are consistent when a larger patch number is applied, as shown in experiment 17-20. 

Another important design choice is the location where the position embedding is added. While many previous methods add it to all encoder layers~\cite{point_bert, point_mae, masked_discrimi}, experiment 11-16 show that it degrades the performance (compared to 5-10). We believe the contradiction is due to the domain gap between datasets. Since position embedding is more informative in point clouds, injecting it into all encoder layers makes the model pay more attention to the key points. Previous works mainly validate their design on small point clouds (\eg, ModelNet40~\cite{3D_ShapeNet_2015_CVPR}). Such behavior might be beneficial in this case since the overall shape is crucial. Nevertheless, for complex point clouds and tasks, the model might neglect fine-grained details. Thus, only injecting patch positions once performs better in our experiments using real-world point clouds. 

\begin{table}[t]
    \centering
    \begin{tabular}{cccccc}
    \toprule
         \textbf{ID} & $r_D$ & $r_M$ & $r_R$ & \textbf{AP25} & \textbf{AP50} \\
    \midrule
         \rowcolor{gray!25}
         1 & 50 & 25 & 25 & \textbf{64.1} & 43.0 \\
         2 & 0 & 90 & 10 & 62.8 & 40.5 \\
    \midrule
         3 & 0 & 75 & 25 & 62.7 & 42.2 \\
         4 & 10 & 65 & 25 & 63.3 & 42.4 \\
         5 & 20 & 55 & 25 & 63.6 & 43.1 \\
         6 & 30 & 45 & 25 & 63.9 & 43.0 \\
         7 & 40 & 35 & 25 & 63.4 & \textbf{44.3} \\
         \rowcolor{gray!25}
           & 50 & 25 & 25 & \textbf{64.1} & 43.0 \\
         8 & 60 & 15 & 25 & 63.8 & 42.4 \\
         9 & 70 & 5 & 25 & 63.2 & 41.2 \\
    \midrule
         10 & 50 & 10 & 40 & 63.6 & 40.2 \\
         11 & 50 & 20 & 30 & 63.8 & 43.2 \\
         \rowcolor{gray!25}
           & 50 & 25 & 25 & \textbf{64.1} & 43.0 \\
         12 & 50 & 30 & 20 & 63.7 & 43.1 \\
         13 & 50 & 40 & 10 & 62.7 & 43.2 \\
    \bottomrule
    \end{tabular}
    \caption{Ablation study on drop patch. $r_D$, $r_M$, $r_R$: the percentage of dropped, masked and reserved patches, respectively. We use FPC and 512 patches. }
    \label{tab:drop_patch}
\end{table}

\noindent
\textbf{Drop Patch.} 
With the benefit of drop patch shown in Table~\ref{tab:det} and~\ref{tab:seg}, we now conduct an ablation study on its hyper-parameters. 
As explained in Section~\ref{subsec:ssp}, drop patch addresses the issue that the position embedding of masked patches makes the MAE pre-training trivial. 
MaskPoint~\cite{masked_discrimi} proposes using an extremely high masked ratio (\SI{90}{\percent}). Experiment 2 and 3 in Table~\ref{tab:drop_patch} show that it does not bring significant improvement because this approach aims to reduce the information leakage caused by overlapping patches. The information leakage due to position embedding is not solved. In experiment 3-9, we fix the ratio of reserved patches and observe the impact of the drop ratio. With only \SI{10}{\percent} percent patches dropped, the model gains an improvement of \SI{0.6}{\percent} AP25 and \SI{0.2}{\percent} AP50. The improvement becomes larger with a higher drop ratio and reaches the maximum at \SI{50}{\percent}. 
A very high drop ratio (\SI{60}{\percent} and \SI{70}{\percent}) is sub-optimal since $r_M$ is low. In this case, the model receives less supervision in the pre-training. 
In experiment 10-13, the drop ratio is fixed. The best performance is achieved when $r_M$ and $r_R$ are approximately equal.

\noindent
\textbf{More Patches \vs More Layers.}
Now we observe the impact of the numbers of encoder layers and patches, with the detection and segmentation head unchanged. The upper half of Table~\ref{tab:layers_patches} shows that more encoder layers harm the performance in object detection. 
Even though the models are pre-trained, only $\sim$78K frames are available for pre-training. 
\begin{table}[bht]
    \centering
    \begin{tabular}{cccccc}
    \toprule
    \textbf{Patches} & \textbf{Layers} & \multicolumn{2}{c}{\textbf{ScanNet Det.}} & \multicolumn{2}{c}{\textbf{S3DIS Seg.}} \\
    &  & \textbf{AP25} & \textbf{AP50} & \textbf{mAcc} & \textbf{mIoU} \\
    \midrule
    512 & 3 & 64.1 & 43.0 & 74.7 & 67.6 \\
    512 & 6 & 63.1 & 42.1 & 76.8 & 70.1 \\
    512 & 12 & 62.1 & 40.7 & \textbf{77.0} & \textbf{70.4} \\
    \midrule
    256 & 3 & 60.8 & 40.4 & 71.5 & 65.0 \\
    1024 & 3 & \textbf{65.6} & \textbf{45.3} & 73.5 & 67.1 \\
    2048 & 3 & 65.0 & 45.2 & 73.6 & 66.7 \\
    \bottomrule
    \end{tabular}
    \caption{Impact of the number of encoder layers and patches. Models are pre-trained using MAE with drop patch}
    \label{tab:layers_patches}
\end{table}
Since the detection head of 3DETR already consists of 8 transformer layers, an encoder with more layers leads to over-fitting. However, adding layers to the encoder improves the performance in segmentation tasks, as the segmentation head is simpler and has fewer parameters. 

The lower half of Table~\ref{tab:layers_patches} shows that using more patches is generally beneficial, as it increases the computation without increasing the number of trainable parameters. However, the effect shows saturation with a large patch number~(\eg, 1024 for detection or 512 for segmentation). 

\begin{table}[thb]
    \centering
    \setlength\tabcolsep{4.2pt}
    \begin{tabular}{lccccc}
         \toprule
         \textbf{Method} & \textbf{Op.} & \textbf{Mem.} & \textbf{Lat.} & \textbf{AP25} & \textbf{AP50} \\
         \midrule
         DPCo & \textbf{5.7} & \textbf{6.6} & 134 & 64.2 & 41.5 \\
         MaskPoint (L3) & 21.4 & 17.3 & 187 & 63.4 & 40.6 \\
         MaskPoint (L12) &  46.9 & 32.0 & 301 & 64.2 & 42.1 \\
         Ours ($M$=512)  & 8.2 & 7.0 & \textbf{73} & 64.1 & 43.0 \\
         Ours ($M$=1024) & 11.7 & 8.7 & 108 & \textbf{65.6} & \textbf{45.3} \\
         \bottomrule
    \end{tabular}
    \caption{Comparison of computational costs in object detection. Op.: Giga floating point operations (GFLOPs). Mem.: memory usage in GB during training with a batch size of 8. Lat.: latency in \si{\ms} is for inference with a batch size of 8 on an NVIDIA Tesla V100. Our models have 3 transformer layers in the backbone. }
    \label{tab:cost_det}
\end{table}

\noindent
\textbf{Computational Costs.} We compare the computational costs of our models with SOTA methods. 
Models in Table~\ref{tab:cost_det} are all pre-trained on ScanNet with self-supervision. 
MaskPoint~\cite{masked_discrimi} uses 2048 patches, following 3DETR.
Our model with 512 patches performs similarly to MaskPoint (L12), while having 5 times lower FLOPs, 4 times less memory usage, and 4 times higher speed, which highlights the efficiency of our model design and the effectiveness of our pre-training. 
Also, the VoteNet pre-trained with DPCo~\cite{li2022closer} is slower than our model because it has more random memory access~\cite{Point_Voxel_CNN_NEURIPS2019}.
When scaled up to 1024 patches, our model achieves significantly higher AP than previous methods with lower costs than MaskPoint (L3). 

More experiments and discussions are available in our supplementary material. 

\section{Conclusion}
In this work, we rethink the application of plain transformers to point clouds. We show that with appropriate designs and self-supervised pre-training, plain transformers are competitive in 3D object detection and semantic segmentation in terms of performance and efficiency. Our work also implies the necessity of evaluating transformers with real-world data, as the designs based on simple and small point clouds might not generalize well. We hope our work can provide a new baseline and inspire more future research on transformers for point cloud understanding. 


{
    \small
    \bibliographystyle{ieeenat_fullname}
    \bibliography{main}
}
\clearpage
\setcounter{page}{1}
\maketitlesupplementary

\appendix

This supplementary material provides additional experimental results, technical details, and visualizations. 
In Section~\ref{app:synthetic}, we evaluate our methods on simple CAD models and explain why we focus on real-world data in the main paper. 
Section~\ref{app:cost} provides more details on the computational costs of our models, including a comparison with the SOTA method PointNeXt.
Section~\ref{app:cl} explains why we use MAE instead of contrastive learning to pre-train the transformers. 
In Section~\ref{app:da}, we discuss the difference between the proposed method drop patch and data augmentation.
Also, we evaluate our models with fewer fine-tuning epochs in Section~\ref{app:fewer}. 
Then, we present more details on our models and training setups in Section~\ref{app:model} and \ref{app:training}, respectively. 
Moreover, we provide the pseudo-code of farthest point clustering in Section~\ref{app:code} and implementation details of this work in Section~\ref{app:implement}.
Finally, we visualize some reconstruction results of a pre-trained MAE in Section~\ref{app:reconst}.

\section{Results on Synthetic Point Clouds}
\label{app:synthetic}
In this work, we focus on large real-world point cloud, while a lot of previous works explore the self-supervised pre-training for transformers on synthetic point clouds. In a standard pipeline, a transformer-based model is pre-trained on ShapeNet~\cite{shapenet_dataset} and then fine-tuned for object classification on ModelNet40~\cite{3D_ShapeNet_2015_CVPR}. In this experiment, we follow this pipeline and compare our results with previous works. 

We follow the standard setup to use a transformer encoder with 12 layers. Each input sample with 1024 points is split into 64 patches. Our models are pre-trained with drop patch. 
As shown in Table~\ref{tab:modelnet_cls}, our model using kNN achieves \SI{93.8}{\percent} overall accuracy on ModelNet40, slightly better than the FPC variant. 
\begin{table}[b]
    \centering
    \begin{tabular}{lc}
    \toprule
        \textbf{Method} & \textbf{OA} (\%) \\
    \midrule
        PointBERT~\cite{point_bert} & 93.2 \\
        POS-BERT~\cite{pos_bert} & 93.6 \\
        PointMAE~\cite{point_mae} & \textbf{93.8} \\
        MaskPoint~\cite{masked_discrimi} & \textbf{93.8} \\
        Ours (FPC) & 93.6 \\
        Ours (kNN) & \textbf{93.8} \\
    \bottomrule
    \end{tabular}
    \caption{Comparison of classfication accuracy on ModelNet40. All models are pre-trained on ShapeNet. OA: overall accuracy on test set. }
    \label{tab:modelnet_cls}
\end{table}
Among previous methods, PointMAE and MaskPoint are most comparable with ours. The former applies a vanilla masked autoencoder for point clouds, whereas the latter addresses the information leakage by learning an implicit function instead of reconstructing the masked patches. Despite different designs, MaskPoint and our method achieve the same accuracy as the plain PointMAE on ModelNet40. On the contrary, they perform differently on real-world data, as shown in Table~\ref{tab:det} in the main paper. 
Thus, we speculate that synthetic point clouds \eg, ShapeNet and ModelNet40, are too simple to reveal the full potential of transformers, so a vanilla MAE (\ie, PointMAE) already reaches the upper bound of the task. The benefit of further optimization is marginal or even unnoticeable. 

This experiment can be viewed as the pilot study of our work. Based on the observation, we believe evaluating transformer-based models on more complex data and tasks is important. Therefore, synthetic single-object point clouds are not the main focus of this work, although they are frequently researched in previous works.

\section{More Discussions on Computational Costs}
\label{app:cost}
\noindent
\textbf{Transformer Layers.} We report the overall GFLOPs (giga floating point operations) of our models in the main paper. In Table~\ref{tab:cost_transformer}, we further show the GFLOPs of transformer layers. Although the multi-head attention has quadratic complexity, the overall GFLOPs of our model only increase from 8.2 to 11.7, when the patch number is doubled (from 512 to 1024). It's because our model has relatively fewer channels and transformer layers. Other components, \eg, patch embedding and feed-forward nets, have a greater impact on the overall costs. 
On the other head, the transformer layers in MaskPoint~\cite{masked_discrimi} with 3 layers have 3.2 times as much GFLOPs as our model with 1024 patches, which is in line with the quadratic complexity.

\begin{table}[b]
    \centering
    \begin{tabular}{lccccc}
    \toprule
        \textbf{Model} & \textbf{Patch} & \textbf{Layer} & \textbf{All} & \textbf{Tr.} & \textbf{AP25} \\
    \midrule
         Ours & 512 & 3 & 8.2 & 3.6 & 64.1 \\
         Ours & 1024 & 3 & 11.7 & 6.7 & 65.6 \\ 
         \midrule
         MaskPoint & 2048 & 3 & 21.4 & 14.1 & 63.4 \\
         MaskPoint & 2048 & 12 & 46.9 & 39.5 & 64.2 \\
    \bottomrule
    \end{tabular}
    \caption{Computational costs of transformer-based object detectors. All: overall GFLOPs. Tr.: GLOPs of transformer layers. }
    \label{tab:cost_transformer}
\end{table}

\noindent
\textbf{Drop Patch.} Besides suppressing the information leakage, our proposed method \emph{drop patch} has the side effect of reducing the computational cost of MAE. With a fixed ratio of reserved patches, the computation in the encoder is unchanged. Meanwhile, the sequence length in the decoder is reduced by the drop ratio, since the decoder ignores the dropped patches. For instance, with 50\% patches dropped, the sequence length of the decoder is halved. 

However, we do not observe a significant training time decrease using drop patch. Despite different drop patch ratios, all models require approximately 8 hours of wall-clock time on a single GPU for pre-training.  
There are two main reasons. First, since the encoder in MAE is heavier than the decoder and cannot be accelerated, the speed up due to drop patch is upper bounded. 
Second, since we convert depth maps into point clouds and perform data augmentation on the flight and the MAE framework has high efficiency~\cite{mae}, our pre-training is bottlenecked by the preprocessing on the CPU. Therefore, decreasing the computation on the GPU has no significant impact on the overall wall-clock time. 

\noindent
\textbf{Semantic Segmentation.}
We report the results on the semantic segmentation task in Table~\ref{tab:cost_seg}. We compare our methods with SOTA PointNeXt~\cite{qian2022pointnext}, a modernized variant of PointNet++~\cite{pointnet++}. 
Our model with 3 encoder layers shows similar performance and throughput as PointNeXt-B. Also, our 12-layer variant achieves higher performance and is more efficient than PointNeXt-L. 
Note that PointNeXt models are not pre-trained but with a more optimized architecture and more inductive bias. The results demonstrate that self-supervised pre-training can close the performance gap between plain transformers and highly optimized models. 

\begin{table}[tb]
    \centering
    \setlength\tabcolsep{4.2pt}
    \begin{tabular}{lccccc}
        \toprule
        \textbf{Method} & \textbf{Op.} & \textbf{Param.} & \textbf{TP} & \textbf{mAcc} &\textbf{mIoU} \\
        \midrule
        PointNeXt-S & \textbf{3.6} & \textbf{0.8} & \textbf{227} & 70.7 & 64.2 \\
        PointNeXt-B & 8.9 & 3.8 & 158 & 74.3 & 67.5 \\
        PointNeXt-L & 15.2 & 7.1 & 115 & 76.1 & 69.5 \\
        \midrule
        Ours (3 layers) & 6.0 & 1.9 & 147 & 74.7 & 67.6 \\
        Ours (6 layers) & 7.2 & 3.5 & 138 & 76.8 & 70.1 \\
        Ours (12 layers) & 9.7 & 6.7 & 123 & \textbf{77.0} & \textbf{70.4} \\
        \bottomrule
    \end{tabular}
    \caption{Computational costs in semantic segmentation task. Same setup as~\cite{qian2022pointnext}. Op.: Giga floating point operations (GFLOPs). Param.: number of parameters in million. TP: throughput during testing in frames per second, with a batch size of 16 on an NVIDIA Tesla V100. Our models use 512 patches. }
    \label{tab:cost_seg}
\end{table}

\section{MAE \vs Contrastive Learning}
\label{app:cl}

\begin{table}[htb]
    \centering
    \setlength\tabcolsep{4.2pt}
    \begin{tabular}{lccc}
        \toprule
        \textbf{Pre-train} & \textbf{Fix Patch Embed} & \textbf{AP25} & \textbf{AP50} \\
        \midrule
        Scratch &   & 61.6 & 38.8 \\
        \midrule
        MoCo & &  57.5 & 38.1 \\
        MoCo & \cmark &  60.0 & 38.4 \\
        MAE & &  62.7 & 42.2 \\
        MAE + Drop Patch & & \textbf{64.1} & \textbf{43.9} \\
        \bottomrule
    \end{tabular}
    \caption{Comparison of MAE and MoCo with point cloud data. Fine-tuning results in object detection on ScanNet dataset. Fix Patch Embed: with patch embedding fixed in pre-training, which stabilizes vision transformers in contrastive learning, proposed in~\cite{chen2021mocov3}. }
    \label{tab:moco}
\end{table}

In this work, we pre-train our models based on the MAE framework. Another possibility is contrastive learning. To compare the two schemes, we pre-train our models using MoCoV3~\cite{mocov2, chen2021mocov3, MOCO_he}, as previous work~\cite{li2022closer} shows that it performs better than methods without negative samples (\eg, BYOL~\cite{BYOL} and SimSiam~\cite{simsiam_Chen_2021_CVPR}) in pre-training with ScanNet data. 
As shown in Table~\ref{tab:moco}, contrastive learning generates worse results than training from scratch. It implies that contrastive learning, which only supervises the global features, is not necessarily beneficial for tasks that require dense features. The observation is also reported in 2D detection~\cite{li2022vitdet, li:2021:benchmarking}. On the contrary, MAE significantly improves the results. Therefore, we use MAE-based pre-training instead of contrastive learning in this work.

\section{Drop Patch \vs Data Augmentation}
\label{app:da}

The drop patch technique is similar to a data augmentation that randomly removes some regions from a point cloud. 
The difference is that drop patch is applied after a point cloud is patchified, so the patchifier is not affected by drop patch. 
On the contrary, if the data augmentation is used instead of drop patch, the input point cloud of the patchifier has ``holes'' in pre-training but does not in fine-tuning. 
Therefore, the generated point patches have different distributions in the two stages, which results in a domain gap between pre-training and fine-tuning. 

\begin{table}[b]
    \centering
    \begin{tabular}{lcc}
         \toprule
         \textbf{Pre-Training} & \textbf{AP25} (\%) & \textbf{AP50} (\%) \\
         \midrule
         From Scratch & 61.6 & 38.8 \\
         \midrule
         MAE & 62.7 & 42.2 \\
         MAE + Drop Patch & \textbf{64.1} & \textbf{43.0} \\
         MAE + Data Aug. & 62.1 & 39.4 \\
         \bottomrule
    \end{tabular}
    \caption{Comparison of drop patch and data augmentation that randomly removes regions from a point cloud.}
    \label{tab:da_drop}
\end{table}

We implement a data augmentation that mimics the behavior of drop patch. 
Assuming the MAE with drop patch uses $M$ patches (before dropping), the data augmentation randomly removes $M/2$ regions with totally \SI{50}{\percent} points from a point cloud to obtain a similar effect. Furthermore, when the data augmentation is applied, the patch number of the MAE is reduced to $M/2$, and the mask ratio is set to \SI{50}{\percent}. 
Also shown in Table~\ref{tab:da_drop}, using data augmentation instead of drop patch significantly worsens the results.

\section{Fune-Tuning with Fewer Epochs}
\label{app:fewer}
\begin{figure}[htb]
    \centering
    \includegraphics[width=0.8\linewidth]{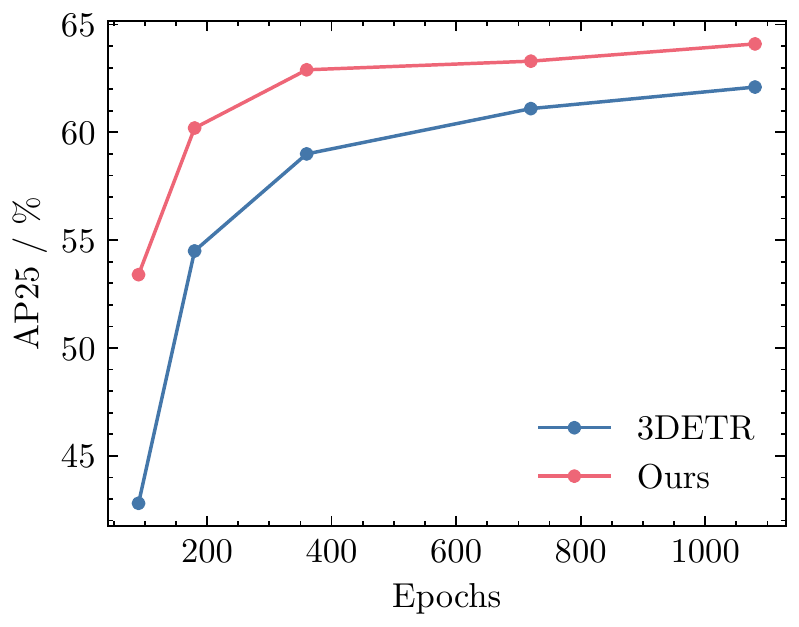}
    \includegraphics[width=0.8\linewidth]{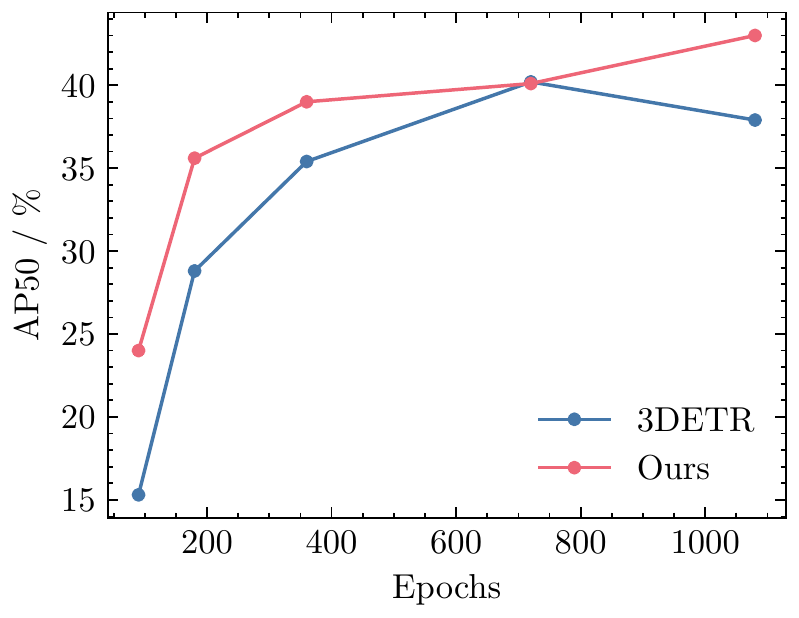}
    \caption{Object detection results on the ScanNet dataset with different epochs. Sample point: 90, 180, 360, 720 and 1080 epochs. }
    \label{fig:epoch_vs_ap}
\end{figure}

In our main paper, we train our 3D detectors for 1080 epochs, following the original setup of 3DETR~\cite{misra2021_3detr}. With the backbone pre-trained, a detector usually converts faster than trained from scratch. In this experiment, we fine-tune our model on the ScanNet detection benchmark with fewer epochs. As shown in Figure~\ref{fig:epoch_vs_ap}, the pre-trained model achieves consistently higher AP25 and AP50 than 3DETR. Moreover, the difference is more significant when the models are trained with fewer epochs. For instance, our detector reaches \SI{10.6}{\percent} higher AP25 and \SI{8.7}{\percent} higher AP50 (absolute) than 3DETR with 90 epochs, whereas the differences are \SI{2.0}{\percent} and \SI{5.1}{\percent} at 1080 epochs. 

We also notice that the pre-trained model still requires a lot of epochs (\ie, 1080) to reach convergence. In 2D object detection, a detector with a backbone pre-trained on ImageNet~\cite{imagenet} usually needs much fewer epochs than trained from scratch. We believe two factors cause the difference. First, our pre-training dataset is much smaller than ImageNet (\ie, 80K samples \vs 1M samples). Also, the 3DETR pipeline is trained with a set-based loss via bipartite matching~\cite{li2022dndetr}, which is known to have a slow convergence~\cite{carion2020detr, li2022dndetr}.

\section{Details of Models}
\label{app:model}
\noindent
\textbf{Patch Embedding.}
We use a PointNet~\cite{pointenet_Qi_2017_CVPR} for patch embedding. The shared MLP consists of 3 layers with 64, 128, and 256 output channels, respectively. The MLP is with Batch Normalization and ReLU activation. Features after the global max-pooling are used as output. 

\noindent
\textbf{Position Embedding.}
Our position embedding consists of two MLPs and a global pooling. The first MLP has 3 layers with 64, 64, and 256 output channels, respectively. The global feature after pooling is concatenated with the 3D coordinates of each key point. The second MLP has 3 layers with 256 channels. All fully connected layers except the last layer are followed by Batch Normalization and ReLU activation. 

\noindent
\textbf{Backbone.}
We use stacked transformer layers as the backbone. The architecture is similar to the backbone in 3DETR~\cite{misra2021_3detr}: each layer has 256 channels, 4 heads, ReLU activation, and a drop out rate of 0.1. The main difference is that we use feed-forward layers with 512 channels instead of 128, as 128 channels lead to under-fitting in our experiments. 

\noindent
\textbf{Pre-Training. }
The MAE decoder consists of 2 transformer layers. Each layer comprises 256 channels, 4 heads, ReLU activation, and a dropout rate of 0.1. The feed-forward sub-network has 256 channels as well. The features from the encoder are projected with a 3-layer MLP, before being fed into the decoder. The approach to reconstructing the masked patches is the same as the original MAE. 

The encoder only encodes reserved patches. Then, the encoder features are appended with a shared learnable masked token, so that the original sequence length is recovered. Also, the position embedding of masked and reserved patches is added to the corresponding features. The decoder then reconstructs masked patches using the sequence as input. 
When drop patch is applied, the dropped patches are neither encoded by the encoder nor reconstructed by the decoder. 
For each point cloud patch, we use a shared linear layer to predict the offset from the key point in the patch to its neighbors. For a patch with $K$ points, the output of the linear layer has $3K$ channels. 

\noindent
\textbf{Semantic Segmentation.}
We first project patch features to 96 channels with two fully connected layers to generate the point-wise semantic prediction. Then, we up-sample the patch features by using nearest neighbor up-sampling~\cite{pointnet++}. For each target coordinate, we search 5 nearest key points. Their distance to the target coordinate is concatenated to their features and projected to 96 channels with an MLP with two layers. Then, we aggregate features by applying a weighted sum according to their inverse distance to the target coordinate. At last, we use an MLP with a drop out rate of 0.5 to perform classification. 

Unlike previous works~\cite{pointnet++, qian2022pointnext}, we do not use any custom CUDA kernels in the segmentation head for simplicity and flexibility. It might have a negative impact on the run-time of our implementation. 

\section{Details of Training Setups}
\label{app:training}
\noindent
\textbf{Pre-Training Data.}
Since the ScanNet dataset consists of registered RGB-D images, we generate point clouds from depth maps on the flight, following~\cite{li2022closer}. We randomly crop depth maps and lift them into 3D space using the camera intrinsic. Then, we randomly sample 20K points from each point cloud. We rotate point clouds to revert the pitch and roll of the camera. We apply horizontal flipping, scaling, and translation. Also, point clouds are randomly rotated around the vertical axis. 
If color channels are applied, we apply Random Contrast and Random Grayscale. We also randomly drop all colors for entire point clouds, following~\cite{qian2022pointnext}

\noindent
\textbf{Hyper-Parameters in Patchifier.}
With our default setup, each point cloud is divided into 512 patches in 3D object detection and semantic segmentation task. 
In the main paper, we also present results with 1024 and 2048 patches. The setups with different patch numbers are summarized in Table~\ref{tab:hyperparam-patch}. Notice that we only apply farthest point clustering for 512 or fewer patches. For 1024 and more patches, we use ball query with a radius of \SI{0.2}{\meter} following previous works~\cite{misra2021_3detr, votenet_Qi_2019_ICCV}. 

\begin{table}[htb]
    \centering
    \begin{tabular}{cccccc}
         \toprule
          \multicolumn{3}{c}{\textbf{Fine-Tune}} & \multicolumn{2}{c}{\textbf{Pre-Train}} & \textbf{Group} \\
         $M$ & $K_\mathrm{det}$ & $K_\mathrm{seg}$ & $M$ & $K$ &  \\
         \midrule
         \textbf{256} & 256 & 128 & 256 & 128 & FPC \\
         \textbf{512} & 128 & 64 & 256 & 128 & FPC \\
         \textbf{1024} & 64 & 64 & 512 & 64 & Ball \\
         \textbf{2048} & 64 & 64 & 1024 & 64 & Ball \\
         \bottomrule
    \end{tabular}
    \caption{Hyper-parameters in patchifier with different patch numbers. $M$: patch number. $K$: sample number in each patch. $K_\mathrm{det}$ and $K_\mathrm{seg}$ are sample numbers per patch for detection and segmentation, respectively. Group: grouping method (farthest point clustering or ball query). }
    \label{tab:hyperparam-patch}
\end{table}

\noindent
\textbf{Evaluation Metrics.} For object detection, we use AP25 and AP50 as metrics. We adopt the evaluation protocol in 3DETR, which reports both metrics when AP25 reaches the maximum. It means the AP25 is the primary metric since AP25 and AP50 might not reach the maximum simultaneously, although they show a similar trend in most cases. 
For semantic segmentation, we report mAcc and mIoU when mIoU reaches the maximum, since mIoU is commonly used as the primary metric in previous works. 

\begin{algorithm}
    \caption{Farthest Point Clustering}
    \label{algo:fpc}
    
    \SetKwInOut{KwInput}{Input}
    \SetKwInOut{KwOut}{Output}

    \KwInput{A point set $\{x_i\}^N_{i=1}$, number of patches $M$,  number of samples in each patch $K$} 
    \KwOut{Assignment matrix $\mathbf{A}_{M \times K}$, $A_{mk} = i$ if $x_i$ is the $k$-th point in the $m$-th patch.}

    \tcc{
        Sample $M$ key points $\{s_i\}^M_{i=1}$ from $\{x_i\}^N_{i=1}$ using Farthest Point Sampling (FPS)
    }
    $\{s_i\}^M_{i=1} \leftarrow \mathrm{FPS}(\{x_i\}^N_{i=1}, M)$

    \tcc{find nearest key point $s_j$ for each point $x_i$}
    \ForEach{$x_i \in \{x_i\}^N_{i=1}$}{
    $t_i \leftarrow \underset{j}{\arg\min}\{ (x_i - s_j)^2 \}$
    }
    \label{algo_line:early_ret}

    Initialize $\mathbf{A}_{M \times K}$ with zeros

    \tcc{make sure each patch has $K$ points}

    \For{$(i\leftarrow 1; \, i \leq M; \, i++)$}{
    \label{algo_line:start_sampling}
    
        $c \leftarrow 0$ \tcp{define a counter}
        
        \For{$( j \leftarrow 1 ; \, j \leq N $ \textbf{and} $ c<K ; \, j++)$}{
            \uIf{$t_j = i$}{
                $A_{ic} \leftarrow j$

                $c ++$
            }
        }

        \tcc{check if the patch has $K$ points}
        $e \leftarrow K - c$ 
        
        \uIf{$e>0$}{
            \tcc{duplicate until the patch has $K$ points}
            \For{$( j \leftarrow 1 ; \, j \leq e ; \, j++)$}{
                $A_{i(c+j)} \leftarrow A_{ij}$
            }
        }
    }
    \label{algo_line:end_sampling}
    \KwRet{$\mathbf{A}_{M \times K}$}
\end{algorithm}

\section{Pseudo-Code for FPC}
\label{app:code}
The pseudo-code of farthest point clustering is shown in Algo.~\ref{algo:fpc}. As mentioned in our main paper, we sample $K$ points in each patch so that each patch has the same number of points (line~\ref{algo_line:start_sampling} to~\ref{algo_line:end_sampling}). The motivation is to make our method more comparable with ball query, since it also applies such sampling. However, this sampling can be omitted if the PointNet patch embedding uses scatter operations\footnote{\eg, \url{https://github.com/rusty1s/pytorch_scatter}}. In this case, the algorithm can directly return the point-wise assignment $t_i$ after line~\ref{algo_line:early_ret}. Although the sampling slightly increases the computation, we empirically find it has no noticeable impact on performance. Thus, we still perform the sampling for better comparability with previous methods.

\section{Implementation Details}
\label{app:implement}
We use PyTorch 1.8.1 with CUDA 10.2 for all training and experiments. 
For pre-training, we generate data based on the code of~\cite{li2022closer}\footnote{\url{https://github.com/lilanxiao/Invar3D}}. For object detection tasks, we modify the open-source code base of 3DETR~\cite{misra2021_3detr}\footnote{\url{https://github.com/facebookresearch/3detr}}. For semantic segmentation, we modify the official code of PointNeXt~\cite{qian2022pointnext}\footnote{\url{https://github.com/guochengqian/PointNeXt}}. 
All speed tests are performed on a cluster node with an Intel Xeon Gold 6230 CPU and 4 NVIDIA Tesla V100 GPUs, each of which has \SI{32}{\giga \byte} memory. However, only one GPU is used for the experiments. 
FLOPs are counted using the open-source library \texttt{fvcore}\footnote{\url{https://github.com/facebookresearch/fvcore}}. 
Our code will be made publicly available.

\section{Reconstruction Results in Pre-Training}
\label{app:reconst}
\begin{figure*}[ht]
    \centering
    \begin{tabular}{c}
        \includegraphics[width=0.8\textwidth]{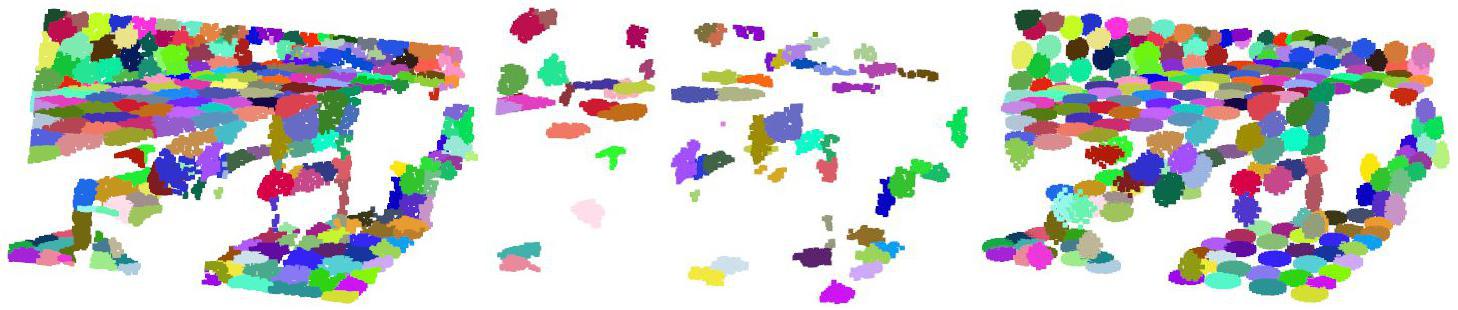} \\
        (a) A long table and some chairs\\
        \includegraphics[width=0.8\textwidth]{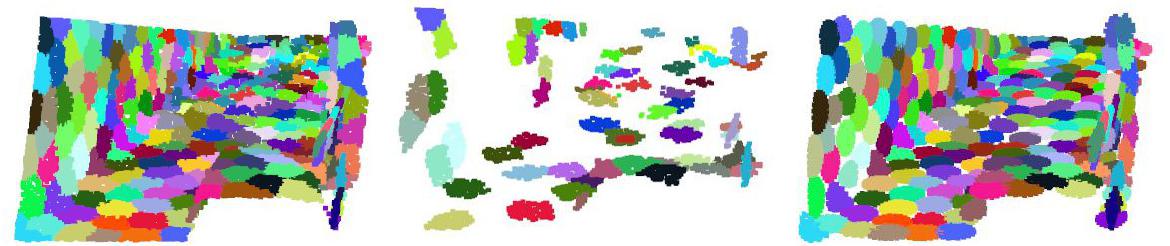} \\
        (b) A corridor \\
        \includegraphics[width=0.8\textwidth]{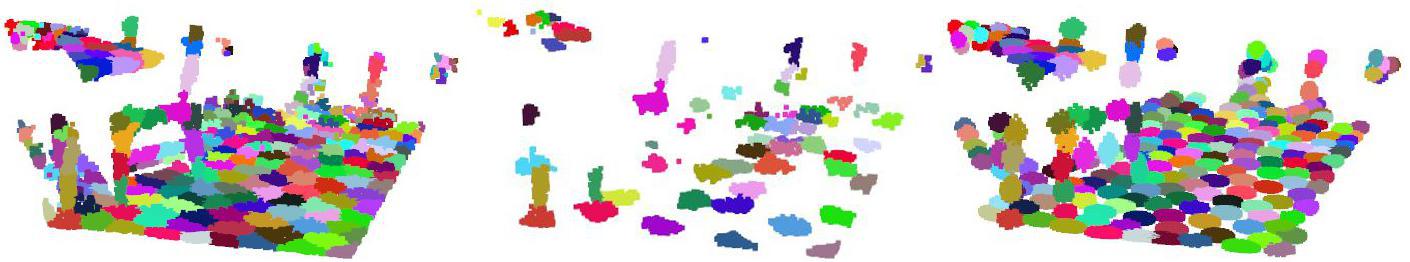} \\
        (c) A table and a chair \\
        \includegraphics[width=0.8\textwidth]{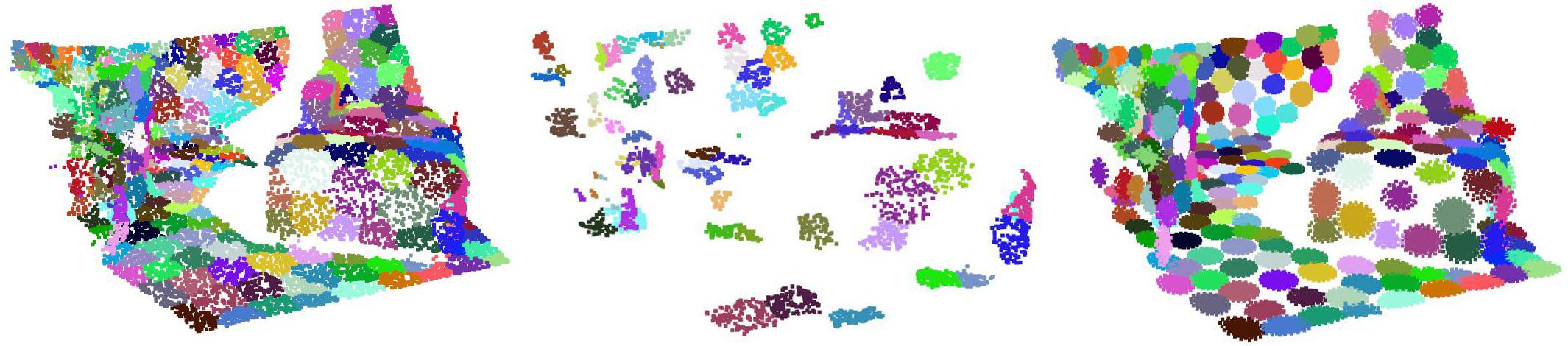} \\
        (d) A bed \\
        \includegraphics[width=0.8\textwidth]{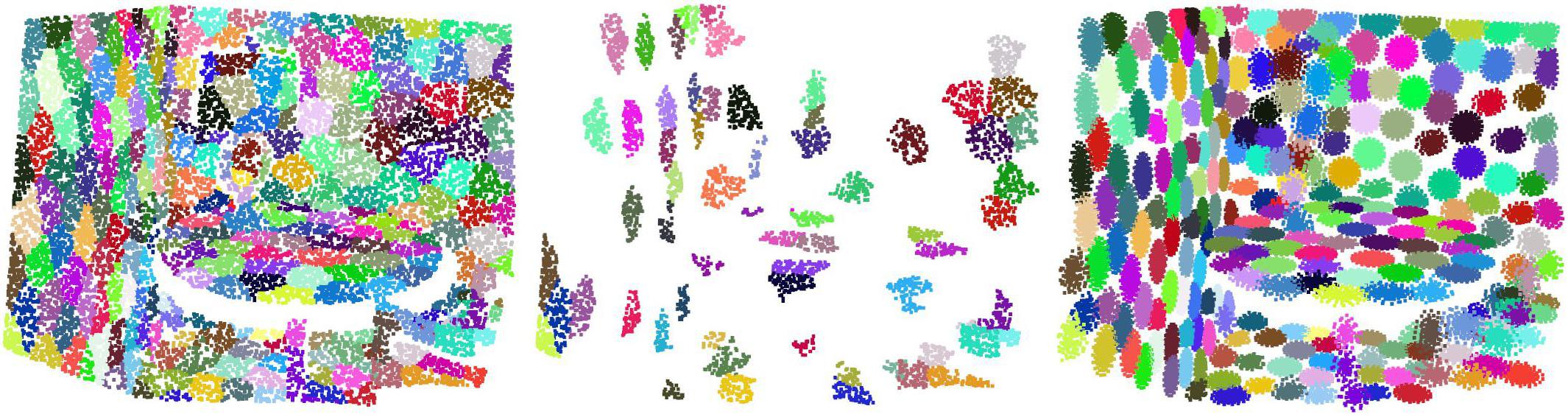} \\
        (e) A round table  
    \end{tabular}
    \caption{Reconstruction results with point clouds. From left to right: original, masked and reconstructed point clouds, respectively. The mask ratio is 75\%. }
    \label{fig:pcd_rec}
\end{figure*}

Here we provide qualitative results of pre-training. We pre-train a model using MAE with drop patch. However, no patch is dropped during the evaluation and the mask ratio is set to \SI{75}{\percent}.
The reconstruction for point clouds is illustrated in Figure~\ref{fig:pcd_rec}. Each patch is painted with a unique color. One can see that MAE mainly reconstructs the low-frequency information of point clouds. Also, the reconstructed patches usually have symmetry, although the ground truth patches have irregular shapes. 

\end{document}